\documentclass[sigconf, screen, table]{acmart}
\usepackage{multirow}
\usepackage{enumitem}

\AtBeginDocument{%
  }

\setcopyright{acmlicensed}
\copyrightyear{2018}
\acmYear{2018}
\acmDOI{XXXXXXX.XXXXXXX}
\acmConference[Conference acronym 'XX]{Make sure to enter the correct
  conference title from your rights confirmation email}{June 03--05,
  2018}{Woodstock, NY}
\acmBooktitle{Woodstock '18: ACM Symposium on Neural Gaze Detection,
 June 03--05, 2018, Woodstock, NY}
\acmISBN{978-1-4503-XXXX-X/2018/06}

\renewcommand\footnotetextcopyrightpermission[1]{}

\begin{document}

\title{DPC-VQA: Decoupling Quality Perception and Residual Calibration for Video Quality Assessment}


%



\author{Xinyue Li}
\authornote{These authors contributed equally to this work.}
\affiliation{%
  \institution{Shanghai Jiao Tong University}
  \city{Shanghai}
  \country{China}
}
\email{xinyueli@sjtu.edu.cn}

\author{Shubo Xu}
\authornotemark[1]
\affiliation{%
  \institution{Baidu Inc.}
  \city{Beijing}
  \country{China}
}
\email{xushubo@baidu.com}

\author{Zhichao Zhang}
\affiliation{%
  \institution{Shanghai Jiao Tong University}
  \city{Shanghai}
  \country{China}
}
\email{liquortect@sjtu.edu.cn}

\author{Zhaolin Cai}
\affiliation{%
  \institution{Xinjiang University}
  \city{Urumqi}
  \country{China}
}
\email{xzm060320@stu.xjtu.edu.cn}

\author{Yitong Chen}
\authornote{Corresponding authors.}
\affiliation{%
  \institution{Shanghai Jiao Tong University}
  \city{Shanghai}
  \country{China}
}
\email{yitongchen@sjtu.edu.cn}

\author{Guangtao Zhai*}
\authornotemark[2]
\affiliation{%
  \institution{Shanghai Jiao Tong University}
  \city{Shanghai}
  \country{China}
}
\email{zhaiguangtao@sjtu.edu.cn}

\renewcommand{\shortauthors}{Xinyue Li et al.}

\begin{abstract}
Recent multimodal large language models (MLLMs) have shown promising performance on video quality assessment (VQA) tasks.
However, adapting them to new scenarios remains expensive due to large-scale retraining and costly mean opinion score (MOS) annotations. 
In this paper, we argue that a pretrained MLLM already provides a useful perceptual prior for VQA, and that the main challenge is to efficiently calibrate this prior to the target MOS space. 
Based on this insight, we propose \textbf{DPC-VQA}, a decoupling perception and calibration framework for video quality assessment. 
Specifically, DPC-VQA uses a frozen MLLM to provide a base quality estimate and perceptual prior, and employs a lightweight calibration branch to predict a residual correction for target-scenario adaptation. 
This design avoids costly end-to-end retraining while maintaining reliable performance with lower training and data costs.
Extensive experiments on both user-generated content (UGC) and AI-generated content (AIGC) benchmarks show that DPC-VQA achieves competitive performance against representative baselines, while using less than 2\% of the trainable parameters of conventional MLLM-based VQA methods and remaining effective with only 20\% of MOS labels. The code will be released upon publication.
\end{abstract}

\ccsdesc[300]{Computing methodologies~Visual inspection}
\ccsdesc[500]{Human-centered computing~Visualization design and evaluation methods}

\keywords{video quality assessment, few-shot, user-generated content, AI-generated content}
\begin{teaserfigure}
  \includegraphics[width=\textwidth]{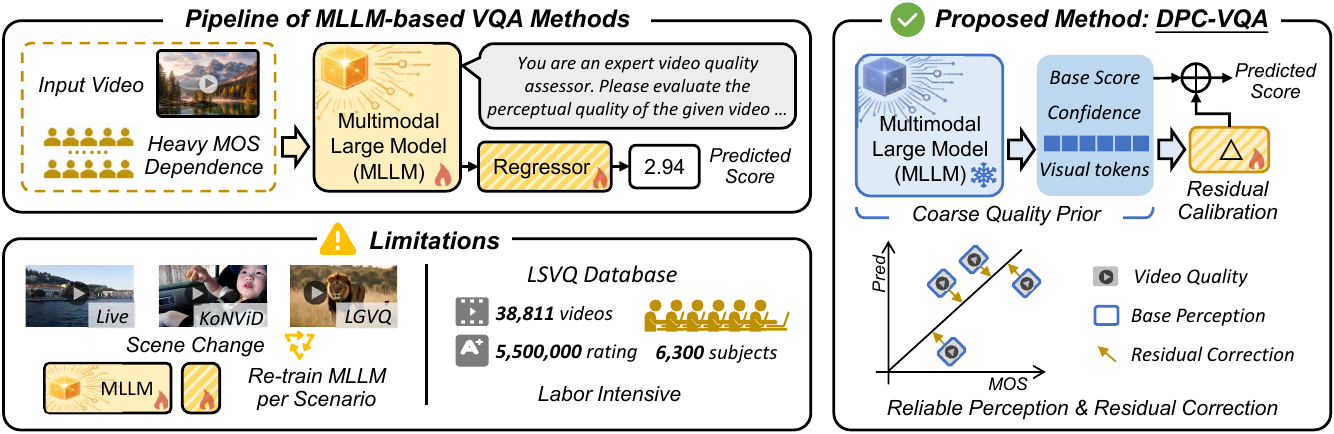}
  \caption{Comparison between existing MLLM-based VQA pipelines and the proposed DPC-VQA framework. Top left: existing methods typically predict quality scores by training or fine-tuning an MLLM with a regressor. Bottom left: when the target scenario changes, they often require scenario-specific retraining with costly MOS supervision. Right: DPC-VQA instead preserves the pretrained MLLM as a frozen quality prior and adapts it through lightweight residual calibration.}
  \label{fig1}
\end{teaserfigure}


\maketitle

\section{Introduction}

\begin{figure*}
\centering
\includegraphics[width=0.99\linewidth]{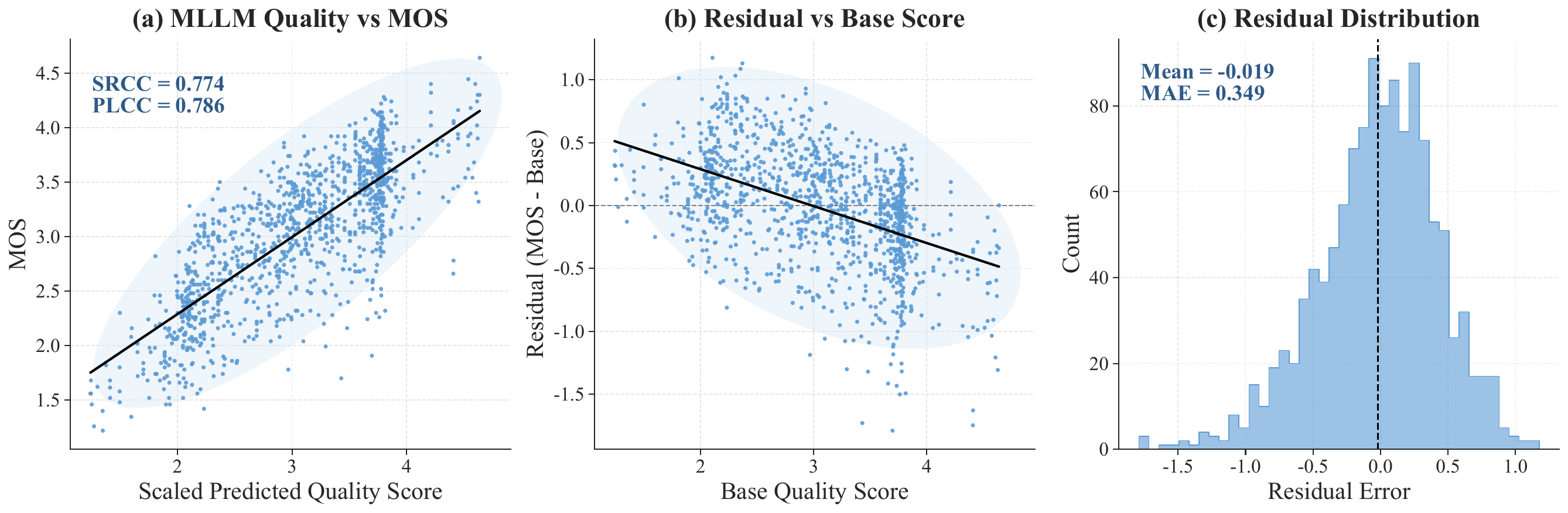}
\caption{
Motivation for decoupling quality perception and residual calibration.
(a) Frozen MLLM quality predictions are already correlated with MOS, indicating a useful base quality prior.
(b) The residual error shows a clear dependency on the base score, suggesting that the remaining discrepancy is structured and learnable.
(c) Residual errors are concentrated around zero, implying that most samples require only limited correction.
These results motivate calibrating a perceptually grounded base score rather than relearning video quality prediction end-to-end. All results are computed on the KoNViD-1k dataset for analysis.
}
\label{fig2}
\end{figure*}

As video content becomes the dominant medium in modern multimedia systems~\cite{PTMVQA,intro1.1,intro1.3}, video quality assessment (VQA) plays an increasingly important role in video acquisition, compression, streaming, transmission, restoration, and generation~\cite{intro2.1,FineVQ,intro2.3}. This broad applicability also requires VQA methods to go beyond strong performance on a single benchmark, and instead support efficient adaptation to diverse real-world scenarios and domains~\cite{UCDA,intro4.3,KVQ,FineVQ}.

Over the past decade, VQA methods have evolved from hand-crafted feature-based models to deep neural networks with increasingly strong representation ability~\cite{PTMVQA,intro6.1,intro6.2}. More recently, multimodal large language models (MLLMs) have further advanced this trend and shown promising performance in video quality assessment~\cite{AIGVA,intro7.1}. These results suggest that pretrained MLLMs can serve as a strong foundation for VQA tasks~\cite{UGVQ,KVQ}.

However, adapting MLLMs to practical VQA scenarios remains difficult~\cite{intro3.1,intro3.2}. Real-world VQA settings vary substantially in content domains, distortion types, capture conditions, and scoring protocols~\cite{intro4.1,UCDA,intro4.3}. When the target scenario changes, existing methods often require retraining or dataset-specific adaptation, which becomes particularly expensive for MLLM-based models due to their large scale~\cite{intro5.1,intro5.2}. At the same time, current VQA pipelines rely on mean opinion score (MOS) annotations, whose collection is notoriously costly because it requires repeated human subjective evaluation~\cite{PTMVQA,li2025iccvgrmpiqa}. These two factors together make large-scale adaptation of MLLM-based VQA difficult in practice, as illustrated in Figure~\ref{fig1}.

In this work, we argue that a pretrained MLLM can already provide a useful generic perceptual prior for VQA. The practical bottleneck is then not how to relearn quality perception from scratch, but how to efficiently calibrate this prior to the MOS space of a target benchmark or scenario. 

Our motivation is supported by three empirical observations. First, as shown in Figure~\ref{fig2}(a), the predictions from a frozen MLLM already exhibit clear correlation with ground-truth MOS, indicating that the pretrained model provides a useful base quality prior. Second, Figure~\ref{fig2}(b) shows that the residual error depends systematically on the base score rather than behaving as random noise, suggesting that the remaining discrepancy is structured and learnable. Third, Figure~\ref{fig2}(c) shows that the residual errors are largely concentrated around zero, implying that most samples require only limited correction rather than full re-estimation. Together, these observations support calibrating a strong pretrained quality prior instead of learning video quality prediction from scratch for each target scenario.

Based on this insight, we decouple VQA into two complementary components: \emph{quality perception} and \emph{score calibration}. We propose \textbf{DPC-VQA}, a \textbf{D}ecoupling \textbf{P}erception and \textbf{C}alibration framework for \textbf{V}ideo \textbf{Q}uality \textbf{A}ssessment. Specifically, the frozen MLLM serves as the perception branch and provides a base quality estimate, while a trainable calibration branch predicts a bounded residual correction in the target MOS space. To improve calibration, the residual predictor can further incorporate auxiliary video features in addition to the frozen MLLM representation. The final prediction is obtained by combining the base score with the learned residual. In this way, the proposed framework leverages the rich quality perception ability already encoded in the pretrained MLLM and adapts it to diverse target scenarios through lightweight calibration rather than costly end-to-end retraining. This design substantially reduces adaptation cost and remains effective even under limited MOS supervision.

Our main contributions are summarized as follows:
\begin{itemize}
    \item We propose \textbf{DPC-VQA}, a decoupling perception and calibration framework for video quality assessment, where a frozen MLLM provides a base quality score and perceptual prior, and a lightweight calibration branch adapts it for target scenarios.
    \item We propose a residual calibration mechanism that refines the base quality prediction via token-wise bounded residual estimation and adaptive aggregation, enabling efficient and fine-grained adaptation to target quality spaces.
    \item Extensive experiments on both UGC and AIGC benchmarks show that the proposed framework achieves competitive performance against representative baselines, while using less than 2\% of the trainable parameters of conventional MLLM-based VQA methods and remaining effective with only 20\% of MOS labels.
\end{itemize}

\section{Related Work}

\label{sec:vqa_related}

\subsection{No-Reference Video Quality Assessment}
\label{sec:nr_vqa}

Video quality assessment (VQA) methods are commonly divided into full-reference (FR), reduced-reference (RR), and no-reference (NR) settings. 
Since reference information is often unavailable in practice, NR-VQA has become the dominant paradigm for perceptual video quality prediction. 
Early NR-VQA methods relied on hand-crafted features and statistical priors. 
For example, TLVQM~\cite{TLVQM} and RAPIQUE~\cite{RAPIQUE} estimate video quality using distortion- and motion-aware descriptors, but their ability to model complex authentic distortions is limited by manually designed features. 
Recent methods instead learn spatiotemporal quality representations directly from MOS-labeled data. 
VSFA~\cite{VSFA} combines spatial features with temporal modeling, while later models such as PVQ~\cite{PatchVQ}, FAST-VQA~\cite{FAST_VQA}, DOVER~\cite{dover}, and SimpleVQA~\cite{simpleVQA} further improve local-to-global modeling, efficiency, and joint spatial-temporal representation learning. 
Despite their strong performance, most existing methods rely heavily on large-scale MOS annotations and are trained end-to-end for specific score spaces, limiting their adaptability under distribution shifts and emerging scenarios.

\subsection{MLLM-based Video Quality Assessment}
\label{sec:mlmm_vqa}

Recent work has begun to adapt large multimodal models (MLLMs) to video quality assessment by reformulating quality prediction as language-conditioned reasoning rather than pure regression. 
LMM-VQA~\cite{LMMVQA} is an early representative model that introduces spatiotemporal visual modeling and instruction tuning for joint quality level prediction and score estimation. 
FineVQ~\cite{FineVQ} extends this direction to fine-grained UGC-VQA, enabling multi-dimensional quality rating, scoring, and attribution. 
CP-LLM~\cite{CPLLM} further emphasizes the complementary roles of global context understanding and pixel-level distortion perception via dual visual encoders, so that the model can output both quality scores and interpretable descriptions. 
More recently, AIGV-Assessor~\cite{AIGVA} targets AI-generated videos by integrating spatiotemporal visual features and prompt information into an LMM, and jointly supports quality-level prediction, score regression, and pairwise comparison. 
Overall, these methods show that MLLMs provide flexible output forms and strong perceptual priors for VQA. 
However, they usually require task-specific adaptation or fine-tuning, and their outputs are not naturally aligned with the MOS scale of a new target domain.

\subsection{Few-shot Video Quality Assessment}
\label{sec:fewshot_vqa}

Few-shot video quality assessment (VQA), which aims to predict MOS in a new domain with only a small number of labeled videos, is practically important because subjective annotation is expensive and time-consuming. 
Unlike mainstream NR-VQA, most few-shot VQA works instead improve label efficiency through self-supervised pretraining, domain adaptation, or pretrained feature reuse.

Representative methods follow this general paradigm. 
CSPT~\cite{CSPT} learns transferable quality-aware representations via contrastive self-supervised pretraining before MOS fine-tuning. 
UCDA~\cite{UCDA} reduces target-domain annotation requirements through unsupervised curriculum domain adaptation across databases. 
PTM-VQA~\cite{PTMVQA} further improves data efficiency by integrating multiple frozen pretrained models in a unified quality-aware space. 
Related label-efficient methods such as CONVIQT~\cite{CONVIQT} and SSL-VQA~\cite{SSL_VQA} also show that self-supervised or semi-supervised representation learning can substantially reduce dependence on large-scale MOS labels.
Overall, existing few-shot VQA studies mainly focus on transferable representation learning and cross-domain adaptation, rather than explicitly modeling few-shot MOS calibration itself.

Overall, prior NR-VQA methods excel at extracting quality cues under fixed training conventions but struggle with cross-dataset score meaning; MLLM-based methods add language-grounded reasoning yet still require reliable conversion to continuous, protocol-consistent scores; and few-shot VQA reduces annotation demand but often entangles adaptation with dataset-specific calibration. These limitations collectively motivate a perception--calibration decoupled approach that separates (i) transferable perceptual evidence extraction from (ii) task-/protocol-specific score calibration, enabling stronger generalization and more controllable deployment across domains and subjective rating protocols.

\section{Method}
\label{sec:method}

\subsection{Problem Formulation and Framework}
\label{sec:formulation_overview}

We study no-reference video quality assessment (VQA), where the goal is to predict a perceptual quality score for an input video.
Given a video $x \in \mathcal{X}$ and its mean opinion score (MOS) $y \in \mathbb{R}$, we learn a predictor $\hat{y}=f(x)$.
MOS values are normalized to $[0,1]$ during training.

We build on the observation that recent MLLMs already provide a useful generic perceptual prior for VQA. 
The main challenge is to efficiently calibrate this prior to the MOS space of a target benchmark or application scenario, while preserving the transferable perception ability of the pretrained model.

Accordingly, we decompose VQA prediction into two components:
\begin{equation}
\hat{y} = q_b + \Delta,
\label{eq:decompose}
\end{equation}
where $q_b$ is a base quality score extracted from a frozen MLLM, and $\Delta$ is a bounded residual calibration term.
This formulation preserves the generic perceptual prior of the MLLM while avoiding expensive retraining for each target scenario.

\begin{figure*}
    \centering
    \includegraphics[width=0.99\linewidth]{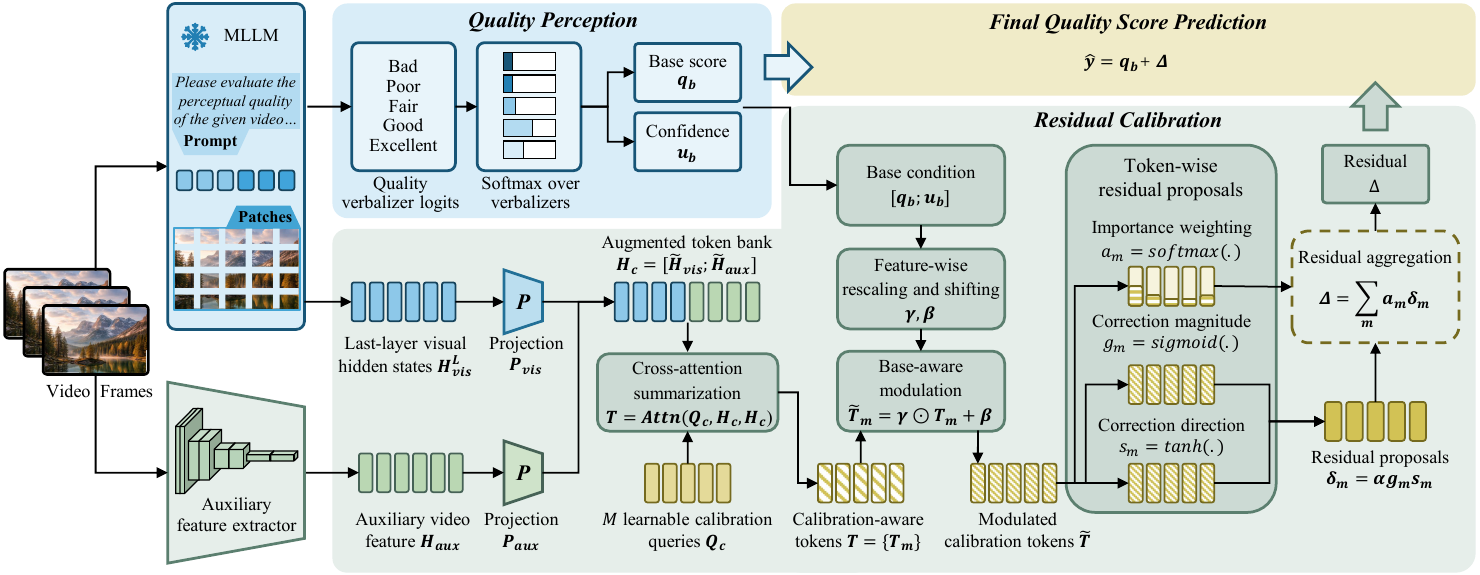}
    \caption{Overview of the proposed DPC-VQA framework. A frozen MLLM first produces a base quality estimate and perceptual representations as a quality prior. A lightweight base-aware residual calibration branch then predicts token-wise corrections conditioned on the base estimate and aggregates them into a residual adjustment. The final quality score is obtained by combining the base estimate with the aggregated residual, enabling efficient adaptation under limited MOS supervision.}
    \label{fig3}
\end{figure*}

The proposed DPC-VQA framework consists of two components: \textbf{quality perception} and \textbf{score calibration}, as shown in Figure~\ref{fig3}.
The perception component provides a generic base quality estimate from a frozen MLLM.
The calibration component predicts the residual discrepancy relative to this base score.

\subsection{Quality Perception}
\label{sec:perception}

Let $\mathcal{M}$ denote a frozen MLLM.
Given an input video $x$, we uniformly sample $T$ frames and feed them into $\mathcal{M}$ together with a fixed quality prompt $p$, which asks the model to answer with a single quality word. The details of the prompts are shown in the supplementary materials.

At the designated answer position, $\mathcal{M}$ produces a next-token logit distribution over its vocabulary.
Meanwhile, from the last transformer block of $\mathcal{M}$, we extract the hidden states corresponding to all visual tokens:
\begin{equation}
H_{\mathrm{vis}}^{L} \in \mathbb{R}^{N \times d_m},
\end{equation}
where $N$ is the number of visual tokens and $d_m$ is the hidden dimension.
The answer-token logits are used to derive a coarse quality score, while $H_{\mathrm{vis}}^{L}$ serves as the frozen visual representation for subsequent residual calibration.

\subsubsection{Quality Verbalizer Distribution}
\label{sec:verbalizer}

We define an ordered set of $K$ quality verbalizers
\begin{equation}
\mathcal{V} = \{v_1, v_2, \dots, v_K\},
\end{equation}
from low to high quality, e.g., \{\emph{bad}, \emph{poor}, \emph{fair}, \emph{good}, \emph{excellent}\}.
To ensure unambiguous extraction, each verbalizer is chosen as a single token in the adopted tokenizer.

Let $z_k$ denote the next-token logit assigned to verbalizer $v_k$.
We compute a restricted softmax over the verbalizer set:
\begin{equation}
p_k = \frac{\exp(z_k)}{\sum_{j=1}^{K}\exp(z_j)},
\qquad k=1,\dots,K.
\label{eq:pk}
\end{equation}
The resulting distribution $\mathbf{p}=[p_1,\dots,p_K]$ represents the MLLM's coarse quality judgment.

\subsubsection{Base Quality Score and Confidence}
\label{sec:base}

Each verbalizer is assigned a scalar anchor $c_k \in [0,1]$ satisfying
\begin{equation}
0 \le c_1 < c_2 < \cdots < c_K \le 1.
\end{equation}
The base quality score is defined as the expectation of the verbalizer distribution:
\begin{equation}
q_b = \sum_{k=1}^{K} p_k c_k.
\label{eq:qb}
\end{equation}
Thus, $q_b$ is directly obtained from the frozen MLLM output without introducing any trainable regression head.

We further estimate the confidence of the base prediction using normalized entropy:
\begin{equation}
u_b = 1 - \frac{-\sum_{k=1}^{K} p_k \log p_k}{\log K},
\label{eq:ub}
\end{equation}
where $u_b \in [0,1]$.
A sharper verbalizer distribution corresponds to a higher confidence.

\subsection{Residual Calibration}
\label{sec:calibration}

While the frozen MLLM provides a useful quality prior, its prediction is not explicitly aligned with the MOS space of a target benchmark.
To model this discrepancy, we learn a lightweight residual calibration module on top of the frozen MLLM representation.
The role of this module is to estimate the residual correction $\Delta$ relative to the base score $q_b$, rather than to replace the base scorer with a new end-to-end predictor.

To make residual estimation more reliable, we further introduce complementary video features extracted by an auxiliary pretrained video encoder.
Let the auxiliary encoder produce a feature sequence
\begin{equation}
H_{\mathrm{aux}} \in \mathbb{R}^{N_a \times d_a},
\end{equation}
where $N_a$ and $d_a$ denote the token number and feature dimension, respectively.
Since $H_{\mathrm{vis}}^{L}$ and $H_{\mathrm{aux}}$ may differ in both dimension and sequence length, we first project them into a shared latent space of dimension $d$:
\begin{equation}
\tilde{H}_{\mathrm{vis}} = P_{\mathrm{vis}}(H_{\mathrm{vis}}^{L}) \in \mathbb{R}^{N \times d},
\qquad
\tilde{H}_{\mathrm{aux}} = P_{\mathrm{aux}}(H_{\mathrm{aux}}) \in \mathbb{R}^{N_a \times d},
\label{eq:proj}
\end{equation}
where $P_{\mathrm{vis}}(\cdot)$ and $P_{\mathrm{aux}}(\cdot)$ are learnable projection layers.

We then form an augmented token bank
\begin{equation}
H_c = [\tilde{H}_{\mathrm{vis}};\tilde{H}_{\mathrm{aux}}]
\in \mathbb{R}^{(N+N_a)\times d}.
\label{eq:hf}
\end{equation}
In this way, the frozen MLLM representation remains the basis of calibration, while the auxiliary features provide complementary evidence for residual estimation.

\subsubsection{Calibration Queries}
\label{sec:queries}
The augmented token bank $H_c$ provides complementary evidence for residual calibration.
To summarize this token sequence into a compact set of calibration-aware representations, we introduce $M$ learnable calibration queries
\begin{equation}
Q_c \in \mathbb{R}^{M \times d},
\end{equation}
and apply cross-attention to the augmented token bank:
\begin{equation}
T = \mathrm{Attn}(Q_c,\; H_c,\; H_c)
\in \mathbb{R}^{M \times d},
\label{eq:T}
\end{equation}
where $T=\{T_1,\dots,T_M\}$.
These query-updated tokens provide a compact representation of residual-relevant evidence for calibration.

\subsubsection{Base-Aware Residual Prediction}
\label{sec:film}

Residual estimation should depend not only on the calibration tokens, but also on the current base judgment.
To encode this dependency, we form a base condition vector
\begin{equation}
c_b = [q_b,\; u_b] \in \mathbb{R}^{2},
\label{eq:cb}
\end{equation}
where $q_b$ denotes the base quality score and $u_b$ denotes its confidence.
Based on $c_b$, we generate feature-wise affine parameters
\begin{equation}
\gamma = W_{\gamma} c_b + b_{\gamma},
\qquad
\beta = W_{\beta} c_b + b_{\beta},
\label{eq:film_param}
\end{equation}
where $\gamma,\beta \in \mathbb{R}^{d}$.
Here, $\gamma$ controls feature-wise rescaling and $\beta$ controls feature-wise shifting.
We then modulate each calibration token by
\begin{equation}
\tilde{T}_m = \gamma \odot T_m + \beta,
\qquad m=1,\dots,M,
\label{eq:film}
\end{equation}
where $\odot$ denotes element-wise multiplication.
In this way, residual estimation is explicitly conditioned on the current base judgment, allowing the calibration module to adapt its correction behavior to different score-confidence states.

Each modulated token $\tilde{T}_m$ predicts a local residual proposal.
Specifically, we estimate its magnitude by
\begin{equation}
g_m = \sigma(w_g^\top \tilde{T}_m + b_g),
\label{eq:gm}
\end{equation}
and its direction by
\begin{equation}
s_m = \tanh(w_s^\top \tilde{T}_m + b_s),
\label{eq:sm}
\end{equation}
where $g_m \in (0,1)$ measures how strongly the current token suggests a correction, and $s_m \in (-1,1)$ determines whether the correction should increase or decrease the base score.
The resulting local residual proposal is
\begin{equation}
\delta_m = \alpha \, g_m \, s_m,
\label{eq:deltam}
\end{equation}
where $\alpha > 0$ is a fixed residual bound.
Therefore, $\delta_m$ can be interpreted as the token-wise correction suggested by the $m$-th calibration token, and each proposal is constrained within $[-\alpha,\alpha]$.

\subsubsection{Residual Aggregation}
\label{sec:aggregation}

Different calibration tokens may contribute unequally to the final correction.
We therefore estimate an importance weight for each token:
\begin{equation}
a_m = \frac{\exp(w_a^\top \tilde{T}_m)}{\sum_{j=1}^{M}\exp(w_a^\top \tilde{T}_j)},
\qquad m=1,\dots,M.
\label{eq:am}
\end{equation}
The final residual is obtained by weighted aggregation:
\begin{equation}
\Delta = \sum_{m=1}^{M} a_m \delta_m.
\label{eq:Delta}
\end{equation}
The final quality prediction is therefore
\begin{equation}
\hat{y} = q_b + \Delta.
\label{eq:yhat}
\end{equation}

\subsection{Training Objective}
\label{sec:loss}

Since the perception branch is fully frozen, training is applied only to the lightweight calibration module, including the projection and residual prediction layers.
The objective is designed to ensure that the trainable branch improves the final score prediction while preserving its role as a lightweight correction to the frozen base judgment.

\subsubsection{Final Score Regression}
\label{sec:loss_reg}

We supervise the final prediction $\hat{y}$ with a Smooth L1 loss:
\begin{equation}
\mathcal{L}_{\mathrm{reg}}
=
\mathrm{SmoothL1}(\hat{y}, y).
\label{eq:Lfinal}
\end{equation}

\subsubsection{Residual Regularization}
\label{sec:loss_res}

To preserve the residual nature of calibration, we further regularize the correction magnitude by
\begin{equation}
\mathcal{L}_{\mathrm{res}}
=
\frac{1}{B}\sum_{i=1}^{B} |\Delta_i|,
\label{eq:Lres}
\end{equation}
where $B$ is the batch size.
This term discourages unnecessarily large corrections and stabilizes training.

\subsubsection{Overall Objective}
\label{sec:loss_total}

The overall training objective is
\begin{equation}
\mathcal{L}
=
\mathcal{L}_{\mathrm{reg}}
+
\lambda_{\mathrm{res}} \mathcal{L}_{\mathrm{res}},
\label{eq:Ltotal}
\end{equation}
where $\lambda_{\mathrm{res}}$ is a balancing coefficient.

\section{Experiments}

\begin{table*}[t]
\centering
\caption{Comparison with representative supervised and few-shot methods on five VQA benchmarks spanning both UGC and AIGC scenarios. Methods marked with $^\ast$ are designed for AIGC-VQA and are therefore reported only on AIGC benchmarks.}
\label{tab_performance}
\resizebox{\textwidth}{!}{
\begin{tabular}{lcccccccccc}
\toprule
\multirow{2}{*}{Methods} & \multicolumn{2}{c}{KoNViD-1k} & \multicolumn{2}{c}{YouTube-UGC} & \multicolumn{2}{c}{LIVE-VQC} & \multicolumn{2}{c}{T2VQA} & \multicolumn{2}{c}{Human-AGVQA} \\
\cmidrule(lr){2-3} \cmidrule(lr){4-5} \cmidrule(lr){6-7} \cmidrule(lr){8-9} \cmidrule(lr){10-11}
& SRCC & PLCC & SRCC & PLCC & SRCC & PLCC & SRCC & PLCC & SRCC & PLCC \\
\midrule
\multicolumn{11}{c}{\textit{supervised}} \\
\midrule
BRISQUE (TIP, 2012)~\cite{BRISQUE}                & 0.634         & 0.615         & 0.472         & 0.447         & 0.459         & 0.451         & 0.505         & 0.507         & 0.438         & 0.446         \\
TLVQM (TIP, 2019)~\cite{TLVQM}                      & 0.773         & 0.769         & 0.669         & 0.659         & 0.658         & 0.639         & 0.610         & 0.613         & 0.582         & 0.584         \\
VSFA (ACMMM, 2019)~\cite{VSFA}                      & 0.794         & 0.799         & 0.787         & 0.789         & 0.718         & 0.771         & 0.704         & 0.708         & 0.618         & 0.633         \\
RAPIQUE (OJSP, 2021)~\cite{RAPIQUE}                 & 0.811         & 0.819         & 0.774         & 0.781         & 0.740         & 0.764         & 0.682         & 0.703         & 0.632         & 0.648         \\
BVQA (TCSVT, 2022)~\cite{BVQA}                      & 0.834         & 0.836         & 0.818         & 0.826         & 0.834         & 0.842         & 0.739         & 0.749         & 0.680         & 0.688         \\
SimpleVQA (ACMMM, 2022)~\cite{simpleVQA}            & 0.856         & 0.860         & 0.847         & 0.856         & 0.845         & 0.859         & 0.628         & 0.639         & 0.695         & 0.709         \\
FastVQA (ECCV, 2022)~\cite{FAST_VQA}                & 0.891         & 0.892         & 0.852         & 0.855         & 0.849         & 0.865         & 0.717         & 0.730         & 0.708         & 0.733         \\
DOVER (ICCV, 2023)~\cite{dover}                     & 0.909         & 0.906         & 0.890         & 0.891         & 0.860         & 0.875         & 0.761         & 0.769         & 0.713         & 0.736         \\
Q-Align (ICML, 2024)~\cite{qalign}                  & 0.865         & 0.877         & 0.831         & 0.847         & 0.773         & 0.829         & 0.760         & 0.777         & 0.756         & 0.763         \\
KVQ (CVPR, 2025)~\cite{KVQ}                         & 0.909         & 0.915         & 0.903         & 0.905         & 0.859         & 0.879         & 0.790         & 0.798         & 0.743         & 0.755         \\
$^\ast$T2VQA (ACMMM, 2024)~\cite{T2VQA}             & -             & -             & -             & -             & -             & -             & 0.796         & 0.806         & 0.751         & 0.754         \\
$^\ast$UGVQ (TOMM, 2025)~\cite{UGVQ}                & -             & -             & -             & -             & -             & -             & 0.803         & 0.811         & 0.748         & 0.757         \\
$^\ast$Human-AGVQA (ACMMM, 2025)~\cite{Human_AGVQA} & -             & -             & -             & -             & -             & -             & 0.808         & 0.813         & 0.780         & 0.788         \\
$^\ast$AIGV-Assessor (CVPR, 2025)~\cite{AIGVA}      & -             & -             & -             & -             & -             & -             & 0.813         & 0.822         & 0.785         & 0.792         \\
\midrule
\multicolumn{11}{c}{\textit{few-shot}} \\
\midrule
UCDA (ICCV, 2021)~\cite{UCDA}                       & 0.785         & 0.791         & 0.756         & 0.761         & 0.762         & 0.770         & 0.468         & 0.481         & 0.462         & 0.481         \\
CSPT (TIP, 2022)~\cite{CSPT}                        & 0.801         & 0.806         & 0.762         & 0.770         & 0.799         & 0.819         & 0.642         & 0.649         & 0.501         & 0.503         \\
VISION (ACMMM, 2022)~\cite{VISION}                  & 0.724         & 0.731         & 0.706         & 0.717         & 0.698         & 0.706         & 0.558         & 0.567         & 0.558         & 0.564         \\
CONTRIQUE (TIP, 2022)~\cite{CONTRIQUE}              & 0.844         & 0.842         & 0.825         & 0.813         & 0.815         & 0.822         & 0.676         & 0.694         & 0.535         & 0.549         \\
CONVIQT (TIP, 2023)~\cite{CONVIQT}                  & 0.851         & 0.849         & 0.832         & 0.822         & 0.808         & 0.817         & 0.653         & 0.661         & 0.594         & 0.607         \\
SSL-VQA (AAAI, 2024)~\cite{SSL_VQA}                 & 0.826         & 0.828         & 0.750         & 0.757         & 0.733         & 0.743         & 0.580         & 0.581         & 0.632         & 0.649         \\
\textbf{Ours}                                       & \textbf{0.883}& \textbf{0.891}& \textbf{0.868}& \textbf{0.871}& \textbf{0.841}& \textbf{0.853}& \textbf{0.782}& \textbf{0.785}& \textbf{0.746}& \textbf{0.753} \\
\bottomrule
\end{tabular}
}
\end{table*}

\subsection{Experimental Setup}
\label{sec:exp_setup}

\subsubsection{Datasets and Evaluation Protocols}
\label{sec:datasets_protocols}

We evaluate the proposed method on five no-reference video quality assessment benchmarks, including three UGC datasets, i.e., KoNViD-1k, YouTube-UGC, and LIVE-VQC, and two AIGC-oriented datasets, i.e., T2VQA and Human-AGVQA.
Following Section~\ref{sec:method}, MOS labels are normalized to $[0,1]$ during training.

Since our setting focuses on few-shot adaptation, we adopt a 5-fold few-shot split protocol.
Each dataset is divided into five non-overlapping folds.
In each run, one fold (20\%) is used as the few-shot training pool, and the remaining four folds (80\%) are used for validation (10\%) and testing (70\%).
Results are averaged over the five runs.

\subsubsection{Evaluation Metrics}
\label{sec:metrics}

We use the Spearman rank-order correlation coefficient (SRCC) and the Pearson linear correlation coefficient (PLCC) for evaluation.
SRCC measures rank consistency between predicted scores and MOS labels, while PLCC measures their linear correlation.
Higher values indicate better agreement with human judgments.
SRCC is treated as the primary metric.

\subsubsection{Implementation Details}
\label{sec:impl_details}

The perception branch is instantiated with a frozen Qwen3-VL-8B model, which produces the base score $q_b$, the confidence score $u_b$, and the last-layer visual hidden states $H_{\mathrm{vis}}^{L}$.
Following Section~\ref{sec:perception}, we uniformly sample $T=4$ frames from each video and feed them into the model with a fixed quality prompt.
The verbalizer set is
$\{\textit{bad}, \textit{poor}, \textit{fair}, \textit{good}, \textit{excellent}\}$.

To improve residual estimation, we introduce auxiliary video features $H_{\mathrm{aux}}$ extracted by SlowFast~\cite{slowfast}.
The MLLM features and auxiliary features are projected into a shared latent space with dimension $d=2048$.
The calibration module uses $M=8$ learnable queries.
The residual bound is set to $\alpha=0.2$, and the residual regularization coefficient in Eq.~(\ref{eq:Ltotal}) is set to $\lambda_{\mathrm{res}}=0.05$.

During training, only the lightweight calibration branch is optimized, while the perception branch remains frozen.
The model is trained for 30 epochs using AdamW with a batch size of 8, a learning rate of $1\times10^{-4}$, and a weight decay of $1\times10^{-4}$.
The training objective consists of the Smooth L1 regression loss in Eq.~(\ref{eq:Lfinal}) and the residual regularization term in Eq.~(\ref{eq:Ltotal}).
The best checkpoint is selected according to the validation SRCC.

\subsubsection{Compared Methods}
\label{sec:compared_methods}

We compare the proposed method with representative VQA approaches from both supervised and few-shot settings, including classical supervised VQA models, recent pretrained-feature-based or transformer-based predictors, few-shot quality adaptation methods, and methods specifically designed for UGC-oriented and AIGC-oriented video quality assessment.
Whenever available, competing methods are evaluated under the same dataset protocol; otherwise, we report the results from the original papers.

\subsection{Comparison with State-of-the-Art Methods}
\label{sec:main_results}

Table~\ref{tab_performance} compares our method with representative supervised and few-shot VQA methods on five benchmarks.
Among the compared few-shot methods, our method achieves the best results on all five benchmarks across both UGC and AIGC scenarios.
This confirms that calibrating a frozen MLLM quality prior is an effective adaptation strategy under limited supervision.

On the three UGC benchmarks, our method obtains 0.883/0.891 on KoNViD-1k, 0.868/0.871 on YouTube-UGC, and 0.841/0.853 on LIVE-VQC in terms of SRCC/PLCC.
Compared with the best few-shot baseline, the SRCC gains are 0.032, 0.036, and 0.026, respectively.
These results show that the proposed residual calibration framework effectively transfers the perceptual prior of a frozen MLLM to authentic distortion scenarios.

On the two AIGC benchmarks, our method also ranks first among few-shot approaches, achieving 0.782/0.785 on T2VQA and 0.746/0.753 on Human-AGVQA.
The SRCC improvements over the best few-shot baseline are 0.106 and 0.114, respectively.
Although some dedicated supervised AIGC-VQA methods achieve higher scores, our method remains competitive with only a few-shot training split and a lightweight calibration branch.

Overall, the results support our main claim that VQA adaptation can be effectively formulated as lightweight residual calibration of a frozen perceptual prior, rather than end-to-end relearning.

\subsection{Ablation Study}
\label{sec:ablation}

Unless otherwise specified, all ablation experiments are conducted on two representative UGC benchmarks, KoNViD-1k and YouTube-UGC.
All variants use the same frozen perception branch, training protocol, and evaluation setting as the full model.
In all cases, the MLLM remains frozen, and only the lightweight calibration branch is optimized. More ablation studies on the MLLM and auxiliary feature extractors are shown in the supplementary materials.

\subsubsection{Residual Calibration versus Direct Regression}
\label{sec:ablation_residual_vs_regression}

We first compare four variants: \emph{Base only}, \emph{Direct regression}, \emph{Score-conditioned regression}, and the proposed \emph{Residual calibration}.
The results are reported in Table~\ref{tab:ablation_core}.

The \emph{Base only} variant already achieves competitive performance, confirming that the frozen MLLM provides a meaningful quality prior.
\emph{Direct regression} further improves the results, indicating that the pretrained features contain useful quality-related information.
Adding $q_b$ as an extra input leads to another consistent gain, showing that the base score is informative even in direct prediction.
However, the proposed residual formulation performs best on both datasets.
This suggests that the improvement does not merely come from reusing $q_b$ as an additional feature; explicitly preserving the frozen perceptual judgment and learning only a lightweight correction is more effective than directly relearning the final score.

\begin{table}[t]
\centering
\caption{Comparison between residual calibration and direct regression on KoNViD-1k and YouTube-UGC.}
\label{tab:ablation_core}
\resizebox{0.49\textwidth}{!}{
\begin{tabular}{lcccc}
\toprule
\multirow{2}{*}{Method} & \multicolumn{2}{c}{KoNViD-1k} & \multicolumn{2}{c}{YouTube-UGC} \\
\cmidrule(lr){2-3} \cmidrule(lr){4-5}
 & SRCC & PLCC & SRCC & PLCC \\
\midrule
Base only ($\hat{y}=q_b$)    & 0.774 & 0.786 & 0.760 & 0.753 \\
Direct regression            & 0.836 & 0.844 & 0.821 & 0.818 \\
Score-conditioned regression & 0.868 & 0.876 & 0.849 & 0.855 \\
Residual calibration (ours)  & \textbf{0.883} & \textbf{0.891} & \textbf{0.868} & \textbf{0.871} \\
\bottomrule
\end{tabular}
}
\end{table}

\subsubsection{Effect of Different Information Sources}
\label{sec:ablation_inputs}

We next study the contributions of different information sources, including the frozen MLLM visual tokens $H_{\mathrm{vis}}^{L}$, the auxiliary video features $H_{\mathrm{aux}}$, the base score $q_b$, and the base confidence $u_b$.
The results are shown in Table~\ref{tab:ablation_inputs}.

Using only $H_{\mathrm{vis}}^{L}$ already yields strong performance, confirming that the frozen MLLM serves as the main perceptual basis of our framework.
By contrast, using only $H_{\mathrm{aux}}$ is clearly weaker, indicating that the auxiliary branch is better viewed as complementary evidence rather than the primary source of quality perception.
Combining $H_{\mathrm{vis}}^{L}$ and $H_{\mathrm{aux}}$ consistently improves performance, showing that temporal or motion-related cues provide useful supplementary information.
Adding $q_b$ brings another gain, suggesting that residual prediction benefits from awareness of the current base judgment.
Finally, incorporating $u_b$ achieves the best results, demonstrating that the confidence of the frozen prediction is also informative for calibration.

\begin{table}[t]
\centering
\caption{Ablation on different information sources used for calibration on KoNViD-1k and YouTube-UGC.}
\label{tab:ablation_inputs}
\begin{tabular}{lcccc}
\toprule
\multirow{2}{*}{Inputs} & \multicolumn{2}{c}{KoNViD-1k} & \multicolumn{2}{c}{YouTube-UGC} \\
\cmidrule(lr){2-3} \cmidrule(lr){4-5}
 & SRCC & PLCC & SRCC & PLCC \\
\midrule
$H_{\mathrm{vis}}^{L}$ only                     & 0.834 & 0.842 & 0.814 & 0.820 \\
$H_{\mathrm{aux}}$ only                         & 0.809 & 0.817 & 0.787 & 0.794 \\
$H_{\mathrm{vis}}^{L}+H_{\mathrm{aux}}$         & 0.854 & 0.862 & 0.836 & 0.841 \\
$H_{\mathrm{vis}}^{L}+H_{\mathrm{aux}}+q_b$     & 0.875 & 0.883 & 0.858 & 0.862 \\
$H_{\mathrm{vis}}^{L}+H_{\mathrm{aux}}+q_b+u_b$ & \textbf{0.883} & \textbf{0.891} & \textbf{0.868} & \textbf{0.871} \\
\bottomrule
\end{tabular}
\end{table}

\subsubsection{Effect of Query-Based Evidence Aggregation}
\label{sec:ablation_queries}

We then examine the aggregation design in the calibration branch by comparing \emph{Mean pooling}, \emph{Self-attention aggregation}, and the proposed \emph{Calibration queries}.
The results are reported in Table~\ref{tab:ablation_queries}.

\emph{Mean pooling} gives the weakest results, while \emph{Self-attention aggregation} performs better by modeling token interactions.
The proposed \emph{Calibration queries} achieve the best performance on both datasets.
These results indicate that residual-relevant cues are not uniformly distributed across tokens, and that explicitly using learnable queries to extract calibration-oriented evidence is more effective than generic sequence aggregation.

We further study the effect of the number of calibration queries $M$ in Table~\ref{tab:ablation_query_num}.
Using only one query is insufficient, while increasing $M$ to $8$ gives the best performance.
Larger values bring no further improvement and may even cause slight degradation, suggesting that a moderate number of calibration queries provides the best trade-off between representation capacity and model simplicity.

\begin{table}[t]
\centering
\caption{Ablation on different evidence aggregation strategies on KoNViD-1k and YouTube-UGC.}
\label{tab:ablation_queries}
\resizebox{0.49\textwidth}{!}{
\begin{tabular}{lcccc}
\toprule
\multirow{2}{*}{Aggregation strategy} & \multicolumn{2}{c}{KoNViD-1k} & \multicolumn{2}{c}{YouTube-UGC} \\
\cmidrule(lr){2-3} \cmidrule(lr){4-5}
 & SRCC & PLCC & SRCC & PLCC \\
\midrule
Mean pooling               & 0.859 & 0.867 & 0.842 & 0.847 \\
Self-attention aggregation & 0.874 & 0.882 & 0.859 & 0.865 \\
Calibration queries (ours) & \textbf{0.883} & \textbf{0.891} & \textbf{0.868} & \textbf{0.871} \\
\bottomrule
\end{tabular}
}
\end{table}

\begin{table}[t]
\centering
\caption{Sensitivity analysis on the number of calibration queries $M$ on KoNViD-1k and YouTube-UGC.}
\label{tab:ablation_query_num}
\setlength{\tabcolsep}{9pt}
\begin{tabular}{ccccc}
\toprule
\multirow{2}{*}{$M$} & \multicolumn{2}{c}{KoNViD-1k} & \multicolumn{2}{c}{YouTube-UGC} \\
\cmidrule(lr){2-3} \cmidrule(lr){4-5}
 & SRCC & PLCC & SRCC & PLCC \\
\midrule
1  & 0.861 & 0.869 & 0.846 & 0.850 \\
4  & 0.874 & 0.882 & 0.858 & 0.863 \\
8  & \textbf{0.883} & \textbf{0.891} & \textbf{0.868} & \textbf{0.871} \\
16 & 0.879 & 0.887 & 0.864 & 0.868 \\
\bottomrule
\end{tabular}
\end{table}

\subsubsection{Effect of Residual Regularization and Residual Bound}
\label{sec:ablation_regularization}

Finally, we study two mechanisms that enforce the residual nature of the calibration branch, namely the residual regularization term and the residual bound.
The results are given in Table~\ref{tab:ablation_lambda} and Table~\ref{tab:ablation_alpha}, respectively.

Table~\ref{tab:ablation_lambda} evaluates the residual regularization coefficient $\lambda_{\mathrm{res}}$ in
\begin{equation}
\mathcal{L}=\mathcal{L}_{\mathrm{reg}}+\lambda_{\mathrm{res}}\mathcal{L}_{\mathrm{res}}.
\end{equation}
Without regularization, the model already performs reasonably well, but introducing a moderate regularization term consistently improves generalization.
The best results are obtained at $\lambda_{\mathrm{res}}=0.05$, while a larger value slightly degrades performance.
This suggests that moderate residual regularization is helpful, whereas overly strong regularization may suppress useful corrections.

Table~\ref{tab:ablation_alpha} further studies the residual bound $\alpha$ in
\begin{equation}
\delta_m = \alpha g_m s_m.
\end{equation}
A very small bound limits the correction capacity, while an overly large bound weakens the calibration prior and makes the model less stable.
The best performance is achieved at $\alpha=0.2$, which supports our design choice that calibration should remain flexible but still be constrained as a limited correction around the frozen base score.

\begin{table}[t]
\centering
\caption{Effect of the residual regularization coefficient $\lambda_{\mathrm{res}}$ on KoNViD-1k and YouTube-UGC.}
\label{tab:ablation_lambda}
\setlength{\tabcolsep}{9pt}
\begin{tabular}{ccccc}
\toprule
\multirow{2}{*}{$\lambda_{\mathrm{res}}$} & \multicolumn{2}{c}{KoNViD-1k} & \multicolumn{2}{c}{YouTube-UGC} \\
\cmidrule(lr){2-3} \cmidrule(lr){4-5}
 & SRCC & PLCC & SRCC & PLCC \\
\midrule
0    & 0.869 & 0.877 & 0.853 & 0.857 \\
0.01 & 0.875 & 0.881 & 0.859 & 0.862 \\
0.05 & \textbf{0.883} & \textbf{0.891} & \textbf{0.868} & \textbf{0.871} \\
0.1  & 0.880 & 0.888 & 0.864 & 0.868 \\
\bottomrule
\end{tabular}
\end{table}

\begin{table}[t]
\centering
\caption{Effect of the residual bound $\alpha$ on KoNViD-1k and YouTube-UGC.}
\label{tab:ablation_alpha}
\setlength{\tabcolsep}{9pt}
\begin{tabular}{ccccc}
\toprule
\multirow{2}{*}{$\alpha$} & \multicolumn{2}{c}{KoNViD-1k} & \multicolumn{2}{c}{YouTube-UGC} \\
\cmidrule(lr){2-3} \cmidrule(lr){4-5}
 & SRCC & PLCC & SRCC & PLCC \\
\midrule
0.1 & 0.871 & 0.879 & 0.856 & 0.860 \\
0.2 & \textbf{0.883} & \textbf{0.891} & \textbf{0.868} & \textbf{0.871} \\
0.3 & 0.876 & 0.884 & 0.860 & 0.865 \\
0.5 & 0.867 & 0.875 & 0.851 & 0.856 \\
\bottomrule
\end{tabular}
\end{table}

\subsection{Efficiency Analysis}
\label{sec:efficiency}

\subsubsection{Data Efficiency}
\label{sec:data_efficiency}

We evaluate data efficiency under limited MOS supervision by varying the training ratio from 5\% to 50\% with respect to the full dataset.
Three variants are compared: direct regression, score-conditioned regression, and the proposed residual calibration, all under the same protocol.

\begin{figure}[t]
\centering
\includegraphics[width=\linewidth]{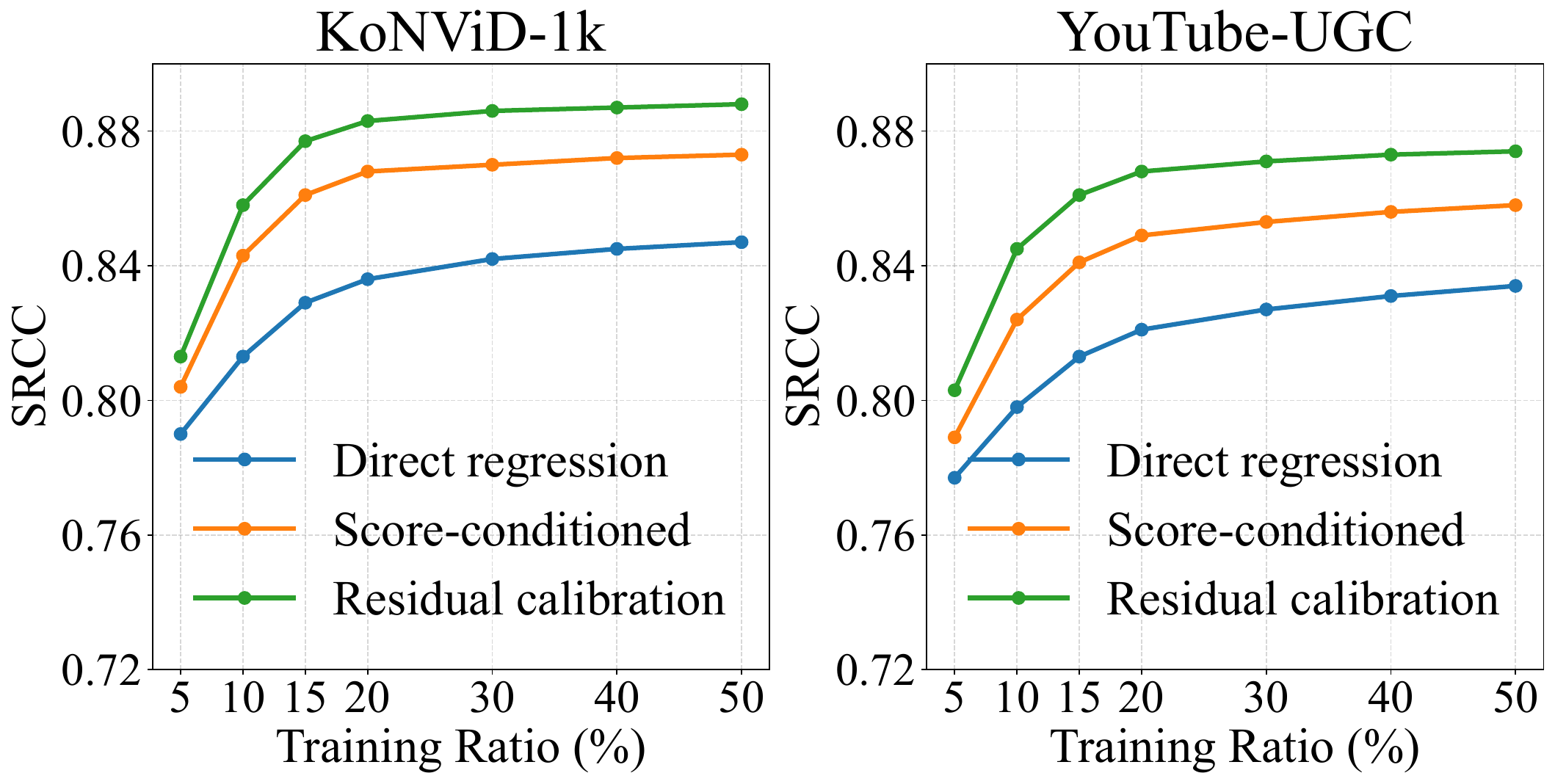}
\caption{Data efficiency comparison under different training ratios on KoNViD-1k and YouTube-UGC.}
\label{fig:data_efficiency}
\end{figure}

Figure~\ref{fig:data_efficiency} shows the SRCC results on KoNViD-1k and YouTube-UGC.
All methods improve as more training data becomes available, but residual calibration consistently performs best across all training ratios.
The gain is especially clear in the low-data regime, indicating that preserving the frozen MLLM quality prior and learning only a residual correction is more effective than directly relearning the MOS mapping.
Notably, strong performance is already achieved with only 10\%--20\% training data, highlighting the practical value of the proposed framework in MOS-expensive scenarios.

\subsubsection{Training Efficiency}
\label{sec:training_efficiency}

We further analyze training efficiency from the optimization perspective.
Although the frozen MLLM still requires forward inference and thus does not reduce GPU memory usage, the proposed method is lightweight in terms of trainable parameters.
Specifically, only the calibration branch is optimized, with 155M trainable parameters, compared with about 8.2B for full fine-tuning.
This corresponds to only about 1.9\% of the trainable parameters of the full model.
In practice, under the default setting, training takes about 2 minutes per epoch on two RTX 4090 GPUs.
These results show that the efficiency advantage of our framework lies in avoiding expensive end-to-end optimization while retaining strong performance.

\section{Conclusion}


In this paper, we revisited MLLM-based no-reference video quality assessment from the perspective of decoupling \emph{quality perception} and \emph{residual calibration}.
Instead of adapting a large model through end-to-end re-training, we treated MLLM as a perceptual prior and formulated adaptation as residual calibration in the target MOS space.
Based on this, we propose \textbf{DPC-VQA}, which uses a frozen MLLM to provide a base quality estimate and perceptual prior, and employs a lightweight calibration branch to predict a residual correction for target-scenario adaptation. 
Experiments on five benchmarks covering both UGC and AIGC scenarios showed that the proposed method achieves the best performance among few-shot methods, while also exhibiting strong data efficiency and low optimization cost.
More broadly, our results indicate that future VQA systems may benefit from shifting the focus from scenario-specific relearning toward efficient calibration of pretrained perceptual priors, opening a more practical path for scalable adaptation under limited MOS supervision.
We hope this perspective can inspire further exploration of efficient adaptation strategies for MLLM-based video quality assessment.

\bibliographystyle{ACM-Reference-Format}
\bibliography{sample-base}

@String{Computing = "Computing" }

@String{Computer = "{IEEE} Computer" }

@String{Springer = "Springer-Verlag" }

@InProceedings{PatchVQ,
    author    = {Ying, Zhenqiang and Mandal, Maniratnam and Ghadiyaram, Deepti and Bovik, Alan},
    title     = {Patch-VQ: 'Patching Up' the Video Quality Problem},
    booktitle = {Proceedings of the IEEE/CVF Conference on Computer Vision and Pattern Recognition (CVPR)},
    month     = {June},
    year      = {2021},
    pages     = {14019-14029}
}

@article{BRISQUE,
  title={No-reference image quality assessment in the spatial domain},
  author={Mittal, Anish and Moorthy, Anush Krishna and Bovik, Alan Conrad},
  journal={IEEE Transactions on image processing},
  volume={21},
  number={12},
  pages={4695--4708},
  year={2012},
  publisher={IEEE}
}

@ARTICLE{BVQA,
  author={Li, Bowen and Zhang, Weixia and Tian, Meng and Zhai, Guangtao and Wang, Xianpei},
  journal={IEEE Transactions on Circuits and Systems for Video Technology}, 
  title={Blindly Assess Quality of In-the-Wild Videos via Quality-Aware Pre-Training and Motion Perception}, 
  year={2022},
  volume={32},
  number={9},
  pages={5944-5958},
  doi={10.1109/TCSVT.2022.3164467}
}

@inproceedings{qalign,
  author       = {Haoning Wu and
                  Zicheng Zhang and
                  Weixia Zhang and
                  Chaofeng Chen and
                  Liang Liao and
                  Chunyi Li and
                  Yixuan Gao and
                  Annan Wang and
                  Erli Zhang and
                  Wenxiu Sun and
                  Qiong Yan and
                  Xiongkuo Min and
                  Guangtao Zhai and
                  Weisi Lin},
  title        = {Q-Align: Teaching LMMs for Visual Scoring via Discrete Text-Defined
                  Levels},
  booktitle    = {Forty-first International Conference on Machine Learning, {ICML} 2024},
  pages        = {54015--54029},
  year         = {2024}
}

@InProceedings{FAST_VQA,
author="Wu, Haoning
and Chen, Chaofeng
and Hou, Jingwen
and Liao, Liang
and Wang, Annan
and Sun, Wenxiu
and Yan, Qiong
and Lin, Weisi",
title="FAST-VQA: Efficient End-to-End Video Quality Assessment with Fragment Sampling",
booktitle="Computer Vision -- ECCV 2022",
year="2022",
publisher="Springer Nature Switzerland",
address="Cham",
pages="538--554",
isbn="978-3-031-20068-7"
}

@INPROCEEDINGS{dover,
  author={Wu, Haoning and Zhang, Erli and Liao, Liang and Chen, Chaofeng and Hou, Jingwen and Wang, Annan and Sun, Wenxiu and Yan, Qiong and Lin, Weisi},
  booktitle={2023 IEEE/CVF International Conference on Computer Vision (ICCV)}, 
  title={Exploring Video Quality Assessment on User Generated Contents from Aesthetic and Technical Perspectives}, 
  year={2023},
  volume={},
  number={},
  pages={20087-20097},
  doi={10.1109/ICCV51070.2023.01843}}

@inproceedings{simpleVQA,
title = {A Deep Learning Based No-Reference Quality Assessment Model for UGC Videos},
author = {Sun, Wei and Min, Xiongkuo and Lu, Wei and Zhai, Guangtao},
booktitle={Proceedings of the 30th ACM International Conference on Multimedia},
year = {2022},
pages = {856–865},
}

@INPROCEEDINGS{SlowFast,
  author={Feichtenhofer, Christoph and Fan, Haoqi and Malik, Jitendra and He, Kaiming},
  booktitle={2019 IEEE/CVF International Conference on Computer Vision (ICCV)}, 
  title={SlowFast Networks for Video Recognition}, 
  year={2019},
  volume={},
  number={},
  pages={6201-6210},
  doi={10.1109/ICCV.2019.00630}
}

@ARTICLE{RAPIQUE,
  author={Tu, Zhengzhong and Yu, Xiangxu and Wang, Yilin and Birkbeck, Neil and Adsumilli, Balu and Bovik, Alan C.},
  journal={IEEE Open Journal of Signal Processing}, 
  title={RAPIQUE: Rapid and Accurate Video Quality Prediction of User Generated Content}, 
  year={2021},
  volume={2},
  number={},
  pages={425-440},
  doi={10.1109/OJSP.2021.3090333}
}

@ARTICLE{TLVQM,
  author={Korhonen, Jari},
  journal={IEEE Transactions on Image Processing}, 
  title={Two-Level Approach for No-Reference Consumer Video Quality Assessment}, 
  year={2019},
  volume={28},
  number={12},
  pages={5923-5938},
  doi={10.1109/TIP.2019.2923051}
}

@inproceedings{VSFA,
author = {Li, Dingquan and Jiang, Tingting and Jiang, Ming},
title = {Quality Assessment of In-the-Wild Videos},
year = {2019},
isbn = {9781450368896},
publisher = {Association for Computing Machinery},
address = {New York, NY, USA},
doi = {10.1145/3343031.3351028},
booktitle = {Proceedings of the 27th ACM International Conference on Multimedia},
pages = {2351–2359},
numpages = {9},
location = {Nice, France},
series = {MM '19}
}

@inproceedings{VISION,
  title={Multiview contrastive learning for completely blind video quality assessment of user generated content},
  author={Mitra, Shankhanil and Soundararajan, Rajiv},
  booktitle={Proceedings of the 30th ACM International Conference on Multimedia},
  pages={1914--1924},
  year={2022}
}

@ARTICLE{CSPT,
  author={Chen, Pengfei and Li, Leida and Wu, Jinjian and Dong, Weisheng and Shi, Guangming},
  journal={IEEE Transactions on Image Processing}, 
  title={Contrastive Self-Supervised Pre-Training for Video Quality Assessment}, 
  year={2022},
  volume={31},
  pages={458-471},
  doi={10.1109/TIP.2021.3130536}
}

@inproceedings{SSL_VQA,
  title={Knowledge guided semi-supervised learning for quality assessment of user generated videos},
  author={Mitra, Shankhanil and Soundararajan, Rajiv},
  booktitle={Proceedings of the AAAI Conference on Artificial Intelligence},
  volume={38},
  number={5},
  pages={4251--4260},
  year={2024}
}

@article{CONVIQT,
  title={Conviqt: Contrastive video quality estimator},
  author={Madhusudana, Pavan C and Birkbeck, Neil and Wang, Yilin and Adsumilli, Balu and Bovik, Alan C},
  journal={IEEE Transactions on Image Processing},
  volume={32},
  pages={5138--5152},
  year={2023},
  publisher={IEEE}
}

@InProceedings{UCDA,
    author    = {Chen, Pengfei and Li, Leida and Wu, Jinjian and Dong, Weisheng and Shi, Guangming},
    title     = {Unsupervised Curriculum Domain Adaptation for No-Reference Video Quality Assessment},
    booktitle = {Proceedings of the IEEE/CVF International Conference on Computer Vision (ICCV)},
    month     = {October},
    year      = {2021},
    pages     = {5178-5187}
}

@article{UGVQ,
author = {Zhang, Zhichao and Sun, Wei and Xinyue, Li and Jia, Jun and Min, Xiongkuo and Zhang, Zicheng and Li, Chunyi and Chen, Zijian and Puyi, Wang and Fengyu, Sun and Shangling, Jui and Zhai, Guangtao},
title = {Benchmarking Multi-dimensional AIGC Video Quality Assessment: A Dataset and Unified Model},
year = {2025},
issue_date = {September 2025},
publisher = {Association for Computing Machinery},
address = {New York, NY, USA},
volume = {21},
number = {9},
issn = {1551-6857},
url = {https://doi.org/10.1145/3749844},
doi = {10.1145/3749844},
journal = {ACM Trans. Multimedia Comput. Commun. Appl.},
month = sep,
articleno = {269},
numpages = {24}
}

@INPROCEEDINGS{AIGVA,
  author={Wang, Jiarui and Duan, Huiyu and Zhai, Guangtao and Wang, Juntong and Min, Xiongkuo},
  booktitle={2025 IEEE/CVF Conference on Computer Vision and Pattern Recognition (CVPR)}, 
  title={AIGV-Assessor: Benchmarking and Evaluating the Perceptual Quality of Text-to-Video Generation with LMM}, 
  year={2025},
  volume={},
  number={},
  pages={18869-18880},
  doi={10.1109/CVPR52734.2025.01758}}

@inproceedings{Human_AGVQA,
  title={Human-activity agv quality assessment: A benchmark dataset and an objective evaluation metric},
  author={Zhang, Zhichao and Sun, Wei and Li, Xinyue and Li, Yunhao and Ge, Qihang and Jia, Jun and Zhang, Zicheng and Ji, Zhongpeng and Sun, Fengyu and Jui, Shangling and others},
  booktitle={Proceedings of the 33rd ACM International Conference on Multimedia},
  pages={6771--6780},
  year={2025}
}

@inproceedings{T2VQA,
  title={Subjective-aligned dataset and metric for text-to-video quality assessment},
  author={Kou, Tengchuan and Liu, Xiaohong and Zhang, Zicheng and Li, Chunyi and Wu, Haoning and Min, Xiongkuo and Zhai, Guangtao and Liu, Ning},
  booktitle={Proceedings of the 32nd ACM International Conference on Multimedia},
  pages={7793--7802},
  year={2024}
}

@ARTICLE{CONTRIQUE,
  author={Madhusudana, Pavan C. and Birkbeck, Neil and Wang, Yilin and Adsumilli, Balu and Bovik, Alan C.},
  journal={IEEE Transactions on Image Processing}, 
  title={Image Quality Assessment Using Contrastive Learning}, 
  year={2022},
  volume={31},
  number={},
  pages={4149-4161},
  doi={10.1109/TIP.2022.3181496}
}

@InProceedings{KVQ,
    author    = {Qu, Yunpeng and Yuan, Kun and Xie, Qizhi and Sun, Ming and Zhou, Chao and Wang, Jian},
    title     = {KVQ: Boosting Video Quality Assessment via Saliency-guided Local Perception},
    booktitle = {Proceedings of the IEEE/CVF Conference on Computer Vision and Pattern Recognition (CVPR)},
    month     = {June},
    year      = {2025},
    pages     = {2150-2160}
}

@ARTICLE{LMMVQA,
  author={Ge, Qihang and Sun, Wei and Zhang, Yu and Li, Yunhao and Ji, Zhongpeng and Sun, Fengyu and Jui, Shangling and Min, Xiongkuo and Zhai, Guangtao},
  journal={IEEE Transactions on Circuits and Systems for Video Technology}, 
  title={LMM-VQA: Advancing Video Quality Assessment With Large Multimodal Models}, 
  year={2025},
  volume={35},
  number={11},
  pages={11083-11096},
  doi={10.1109/TCSVT.2025.3571788}}

@inproceedings{li2025iccvgrmpiqa,
  title        = {Few-Shot Image Quality Assessment via Adaptation of Vision-Language Models},
  author       = {Li, Xudong and Huang, Zihao and Zhang, Yan and Shen, Yunhang and Li, Ke and Zheng, Xiawu and Cao, Liujuan and Ji, Rongrong},
  booktitle    = {Proceedings of the IEEE/CVF International Conference on Computer Vision (ICCV)},
  year         = {2025},
  month        = oct,
  pages        = {10442--10452}
}

@inproceedings{PTMVQA,
  author    = {Kun Yuan and Hongbo Liu and Mading Li and Muyi Sun and Ming Sun and Jiachao Gong and Jinhua Hao and Chao Zhou and Yansong Tang},
  title     = {PTM-VQA: Efficient Video Quality Assessment Leveraging Diverse PreTrained Models from the Wild},
  booktitle = {Proceedings of the IEEE/CVF Conference on Computer Vision and Pattern Recognition (CVPR)},
  pages     = {2835--2845},
  year      = {2024},
  doi       = {10.1109/CVPR52733.2024.00274}
}

@article{CPLLM,
  title   = {CP-LLM: Context and Pixel Aware Large Language Model for Video Quality Assessment},
  author  = {Wen, Wen and Wu, Yaohong and Sheng, Yue and Birkbeck, Neil and Adsumilli, Balu and Wang, Yilin},
  journal = {CoRR},
  volume  = {abs/2505.16025},
  year    = {2025},
  doi     = {10.48550/arXiv.2505.16025}
}

@inproceedings{FineVQ,
  title     = {FineVQ: Fine-Grained User Generated Content Video Quality Assessment},
  author    = {Duan, Huiyu and Hu, Qiang and Wang, Jiarui and Yang, Liu and Xu, Zitong and Liu, Lu and Min, Xiongkuo and Cai, Chunlei and Ye, Tianxiao and Zhang, Xiaoyun and Zhai, Guangtao},
  booktitle = {Proceedings of the IEEE/CVF Conference on Computer Vision and Pattern Recognition},
  pages     = {3206--3217},
  year      = {2025},
  doi       = {10.1109/CVPR52734.2025.00305}
}

@INPROCEEDINGS{intro1.1,
  author={Conde, Marcos V. and Zadtootaghaj, Saman and Barman, Nabajeet and Timofte, Radu and He, Chenlong and Zheng, Qi and Zhu, Ruoxi and Tu, Zhengzhong and Wang, Haiqiang and Chen, Xiangguang and Meng, Wenhui and Pan, Xiang and Shi, Huiying and Zhu, Han and Xu, Xiaozhong and Sun, Lei and Chen, Zhenzhong and Liu, Shan and Zhang, Zicheng and Wu, Haoning and Zhou, Yingjie and Li, Chunyi and Liu, Xiaohong and Lin, Weisi and Zhai, Guangtao and Sun, Wei and Cao, Yuqin and Jiang, Yanwei and Jia, Jun and Zhang, Zhichao and Chen, Zijian and Zhang, Weixia and Min, Xiongkuo and Göring, Steve and Qi, Zihao and Feng, Chen},
  booktitle={2024 IEEE/CVF Conference on Computer Vision and Pattern Recognition Workshops (CVPRW)}, 
  title={AIS 2024 Challenge on Video Quality Assessment of User-Generated Content: Methods and Results}, 
  year={2024},
  volume={},
  number={},
  pages={5826-5837},
  doi={10.1109/CVPRW63382.2024.00591}}

@misc{intro1.3,
      title={Video Quality Assessment: A Comprehensive Survey}, 
      author={Qi Zheng and Yibo Fan and Leilei Huang and Tianyu Zhu and Jiaming Liu and Zhijian Hao and Shuo Xing and Chia-Ju Chen and Xiongkuo Min and Alan C. Bovik and Zhengzhong Tu},
      year={2024},
      eprint={2412.04508},
      archivePrefix={arXiv},
      primaryClass={eess.IV},
      url={https://arxiv.org/abs/2412.04508}, 
}

@InProceedings{intro2.1,
    author    = {Qi, Zelu and Shi, Ping and Zhang, Chaoyang and Wang, Shuqi and Zhao, Fei and Pan, Da and Ying, Zefeng},
    title     = {Towards Holistic Visual Quality Assessment of AI-Generated Videos: A LLM-Based Multi-Dimensional Evaluation Model},
    booktitle = {Proceedings of the IEEE/CVF Conference on Computer Vision and Pattern Recognition (CVPR) Workshops},
    month     = {June},
    year      = {2025},
    pages     = {1493-1502}
}

@InProceedings{intro2.3,
    author    = {Zhu, Hanwei and Wu, Haoning and Zhang, Zicheng and Zhu, Lingyu and Li, Yixuan and Chen, Peilin and Wang, Shiqi and Zhou, Chris Wei and Cao, Linhan and Sun, Wei and Zhu, Xiangyang and Zhang, Weixia and Zhu, Yucheng and Liu, Jing and Zhu, Dandan and Zhai, Guangtao and Min, Xiongkuo and Zhang, Zhichao and Li, Xinyue and Xu, Shubo and Dao, Anh and Li, Yifan and Yu, Hongyuan and Yi, Jiaojiao and Tian, Yiding and Wu, Yupeng and Sun, Feiran and Jiao, Lijuan and Jiang, Song},
    title     = {VQualA 2025 Challenge on Visual Quality Comparison for Large Multimodal Models: Methods and Results},
    booktitle = {Proceedings of the IEEE/CVF International Conference on Computer Vision (ICCV) Workshops},
    month     = {October},
    year      = {2025},
    pages     = {3383-3393}
}

@InProceedings{intro3.1,
    author    = {Zanella, Maxime and Fuchs, Cl\'ement and De Vleeschouwer, Christophe and Ben Ayed, Ismail},
    title     = {Realistic Test-Time Adaptation of Vision-Language Models},
    booktitle = {Proceedings of the IEEE/CVF Conference on Computer Vision and Pattern Recognition (CVPR)},
    month     = {June},
    year      = {2025},
    pages     = {25103-25112}
}

@InProceedings{intro3.2,
    author    = {Zhang, Xingxuan and Li, Jiansheng and Chu, Wenjing and hai, junjia and Xu, Renzhe and Yang, Yuqing and Guan, Shikai and Xu, Jiazheng and Jing, Liping and Cui, Peng},
    title     = {On the Out-Of-Distribution Generalization of Large Multimodal Models},
    booktitle = {Proceedings of the IEEE/CVF Conference on Computer Vision and Pattern Recognition (CVPR)},
    month     = {June},
    year      = {2025},
    pages     = {10315-10326}
}

@InProceedings{intro4.1,
    author    = {Zhang, Weixia and Zheng, Bingkun and Chen, Junlin and Wang, Zhihua},
    title     = {Multi-Dimensional Quality Assessment for UGC Videos via Modular Multi-Modal Vision-Language Models},
    booktitle = {Proceedings of the IEEE/CVF Conference on Computer Vision and Pattern Recognition (CVPR) Workshops},
    month     = {June},
    year      = {2025},
    pages     = {1557-1566}
}

@InProceedings{intro4.3,
    author    = {Wen, Wen and Li, Mu and Zhang, Yabin and Liao, Yiting and Li, Junlin and Zhang, Li and Ma, Kede},
    title     = {Modular Blind Video Quality Assessment},
    booktitle = {Proceedings of the IEEE/CVF Conference on Computer Vision and Pattern Recognition (CVPR)},
    month     = {June},
    year      = {2024},
    pages     = {2763-2772}
}

@InProceedings{intro5.1,
    author    = {Xia, Jiaer and Tong, Bingkui and Zang, Yuhang and Shao, Rui and Zhou, Kaiyang},
    title     = {Bootstrapping Grounded Chain-of-Thought in Multimodal LLMs for Data-Efficient Model Adaptation},
    booktitle = {Proceedings of the IEEE/CVF International Conference on Computer Vision (ICCV)},
    month     = {October},
    year      = {2025},
    pages     = {208-217}
}

@InProceedings{intro5.2,
    author    = {Liang, Tian and Huang, Jing and Kong, Ming and Chen, Luyuan and Zhu, Qiang},
    title     = {Querying as Prompt: Parameter-Efficient Learning for Multimodal Language Model},
    booktitle = {Proceedings of the IEEE/CVF Conference on Computer Vision and Pattern Recognition (CVPR)},
    month     = {June},
    year      = {2024},
    pages     = {26855-26865}
}

@misc{intro6.1,
      title={DCVQE: A Hierarchical Transformer for Video Quality Assessment}, 
      author={Zutong Li and Lei Yang},
      year={2022},
      eprint={2210.04377},
      archivePrefix={arXiv},
      primaryClass={cs.CV},
      url={https://arxiv.org/abs/2210.04377}, 
}

@INPROCEEDINGS{intro6.2,
  author={Wang, Yilin and Ke, Junjie and Talebi, Hossein and Yim, Joong Gon and Birkbeck, Neil and Adsumilli, Balu and Milanfar, Peyman and Yang, Feng},
  booktitle={2021 IEEE/CVF Conference on Computer Vision and Pattern Recognition (CVPR)}, 
  title={Rich features for perceptual quality assessment of UGC videos}, 
  year={2021},
  volume={},
  number={},
  pages={13430-13439},
  doi={10.1109/CVPR46437.2021.01323}}

@INPROCEEDINGS{intro7.1,
  author={Zhang, Zicheng and Jia, Ziheng and Wu, Haoning and Li, Chunyi and Chen, Zijian and Zhou, Yingjie and Sun, Wei and Liu, Xiaohong and Min, Xiongkuo and Lin, Weisi and Zhai, Guangtao},
  booktitle={2025 IEEE/CVF Conference on Computer Vision and Pattern Recognition (CVPR)}, 
  title={Q-Bench-Video: Benchmark the Video Quality Understanding of LMMs}, 
  year={2025},
  volume={},
  number={},
  pages={3229-3239},
  doi={10.1109/CVPR52734.2025.00307}}



\end{document}